\documentclass{article}
\usepackage{spconf}
\usepackage{times}
\usepackage{epsfig}
\usepackage{graphicx}
\usepackage{amsmath}
\usepackage{amssymb}
\usepackage{enumerate}
\usepackage{enumitem}
\usepackage{bm}
\usepackage{amsmath}
\usepackage{amssymb}
\usepackage{amsfonts}
\usepackage{multirow}
\usepackage{tabularx}
\usepackage{mathrsfs}
\usepackage{amsthm}
\usepackage{booktabs}

\newcommand{\xiaok}[1]{\left(#1\right)}

\newcommand{\argmax}[1]{{\mathop{\arg\mathrm{max}}_{#1}\,}}

\newcommand{\opseq}[3]{{#1_1 #3 #1_2 #3 \cdots #3 #1_{#2}}}
\newcommand{\seq}[2]{\opseq{#1}{#2}{,}}

\newcommand{\bmc}{\bm{c}}

\newcommand{\bmz}{\bm{z}}

\newcommand{\bmC}{\bm{C}}

\newcommand{\calC}{\mathcal{C}}

\newcommand{\calL}{\mathcal{L}}

\newcommand{\bbE}{\mathbb{E}}

\newcommand{\bbR}{\mathbb{R}}

\usepackage[pagebackref,breaklinks,colorlinks,citecolor=blue]{hyperref}
\usepackage[capitalize]{cleveref}
\crefname{section}{Sec.}{Secs.}
\Crefname{section}{Section}{Sections}
\Crefname{table}{Table}{Tables}
\crefname{table}{Tab.}{Tabs.}

\title{Transformer-based Clipped Contrastive Quantization Learning for Unsupervised Image Retrieval}

\name{Ayush Dubey$^{\ast}$, Shiv Ram Dubey$^{\ast}$, Satish Kumar Singh$^{\ast}$, Wei-Ta Chu$^{\dagger}$}
\address{$^{\ast}$Computer Vision and Biometrics Lab, Indian Institute of Information Technology Allahabad, India\\
$^{\dagger}$Department of Computer Science and Information Engineering, National Cheng Kung University, Taiwan\\
{\tt\small mit2021040@iiita.ac.in, srdubey@iiita.ac.in, sk.singh@iiita.ac.in, wtchu@gs.ncku.edu.tw}
}

\begin{document}

\maketitle

\begin{abstract}
Unsupervised image retrieval aims to learn the important visual characteristics without any given level to retrieve the similar images for a given query image. The Convolutional Neural Network (CNN)-based approaches have been extensively exploited with self-supervised contrastive learning for image hashing. However, the existing approaches suffer due to lack of effective utilization of global features by CNNs and biased-ness created by false negative pairs in the contrastive learning. In this paper, we propose a TransClippedCLR model by encoding the global context of an image using Transformer having local context through patch based processing, by generating the hash codes through product quantization and by avoiding the potential false negative pairs through clipped contrastive learning. The proposed model is tested with superior performance for unsupervised image retrieval on benchmark datasets, including CIFAR10, NUS-Wide and Flickr25K, as compared to the recent state-of-the-art deep models. The results using the proposed clipped contrastive learning are greatly improved on all datasets as compared to same backbone network with vanilla contrastive learning. 
\end{abstract}


\section{Introduction}

\begin{figure}[t]
  \centering
  \includegraphics[trim={10 80 10 0},clip, width=0.9\columnwidth]{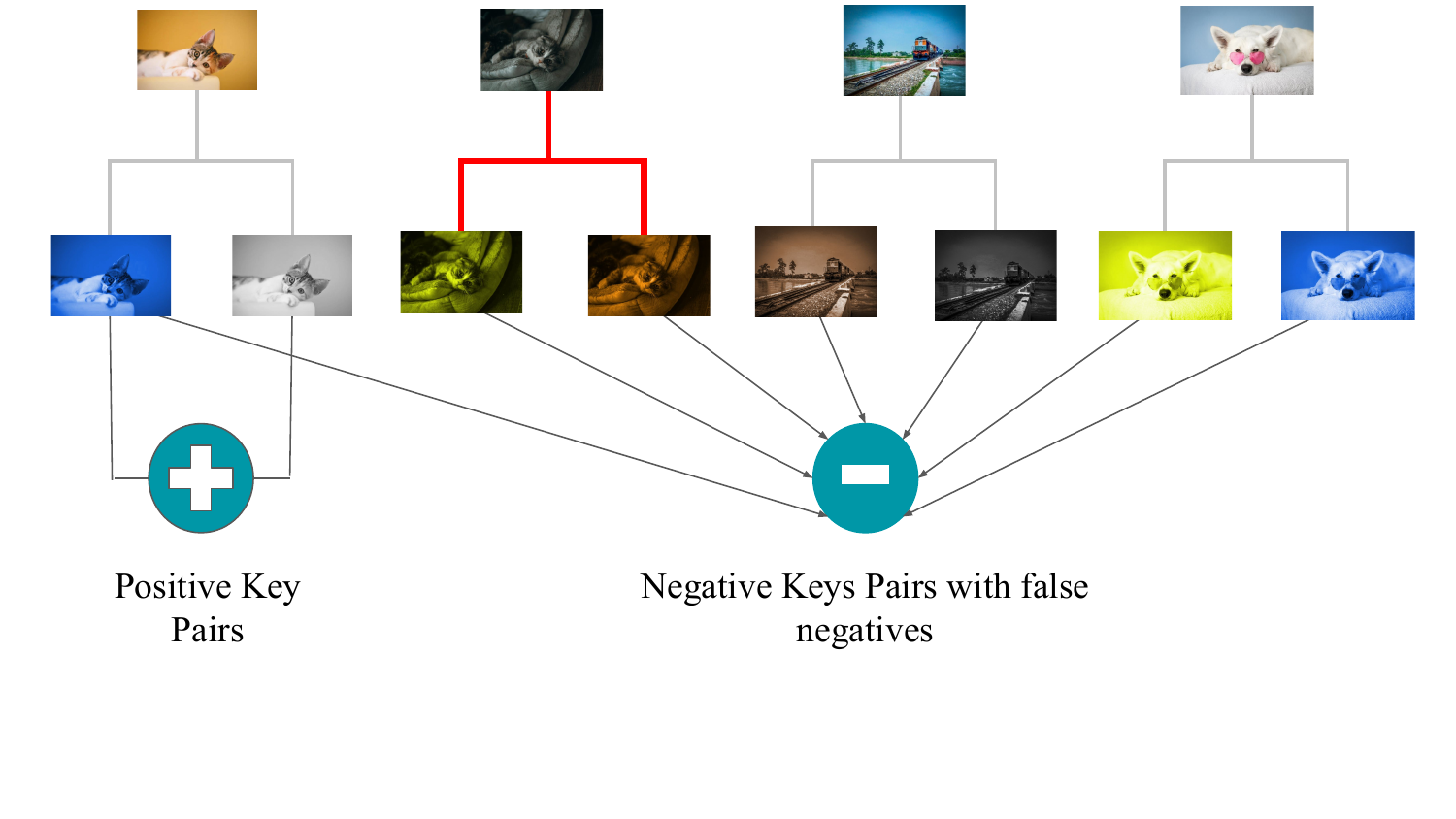}
    \vspace{-3mm}
  \caption{The augmented view is considered as the positive key while rest of the views from batch make up the negative keys. Because the negative keys are randomly sampled false negatives can occur and hamper training. The views of the second image of the cat marked in red are false negatives.}
\label{fig:motivation}
\end{figure}

With the advent of deep learning, a huge progress is made for content-based image retrieval using CNN-based approaches \cite{dubey2021decade}, including supervised \cite{csq}, \cite{dubey2022vision} and unsupervised \cite{deepbit}, \cite{mecoq} models. The supervised image retrieval requires the large-scale annotated datasets. Hence, the unsupervised image hashing and retrieval is the demand for large-scale image retrieval systems, where getting a labeled dataset is very challenging, time-consuming and laborious. 
Some common approaches utilized for unsupervised image retrieval include self-supervision, data augmentation, contrastive learning, product quantization, learning a codebook, and energy optimization \cite{dubey2021decade}. 
The unsupervised hashing approaches using deep learning usually follow either deep binary hashing (DBH) \cite{deepbit} or deep quantization (DQ) \cite{spq}, \cite{mecoq}. The minimal quantization loss, evenly distributed codes and uncorrelated bits are the criterion utilized by DeepBit \cite{deepbit}. The prominent DBH methods include 
Semantic Structure based unsupervised Deep Hashing (SSDH) \cite{ssdh}, 
Twin-Bottleneck Hashing (TBH) \cite{tbh}, 
BiHalfNet \cite{bihalfnet}, 
Contrastive Information Bottleneck Hashing (CIBHash) \cite{cibhash}, etc.
The DBH methods use a CNN to extract the real-valued features which are converted into the binary equivalent. However, such approaches usually suffer due to diminishing gradient problems for hard quantization.
Deep quantization based methods \cite{spq}, \cite{mecoq} involve compressing the weights of the model and mapping them to a smaller set of discrete values, which are then used for image feature extraction. DBD-MQ considers binarization as a multi-quantization task and uses a K-AutoEncoders network to jointly learn the parameters and binarization functions \cite{dbd-mq}. This allows for discriminative binary descriptors with fine-grained multi-quantization \cite{dbd-mq}. DeepQuan learns the binary codes by exploiting the product quantization (PQ) by minimizing the quantization error. It utilizes a manifold preserving loss and a weighted triplet loss for training \cite{deepquan}.

\begin{figure*}[t]
  \centering
  \includegraphics[width=0.8\textwidth, height=6.0cm]{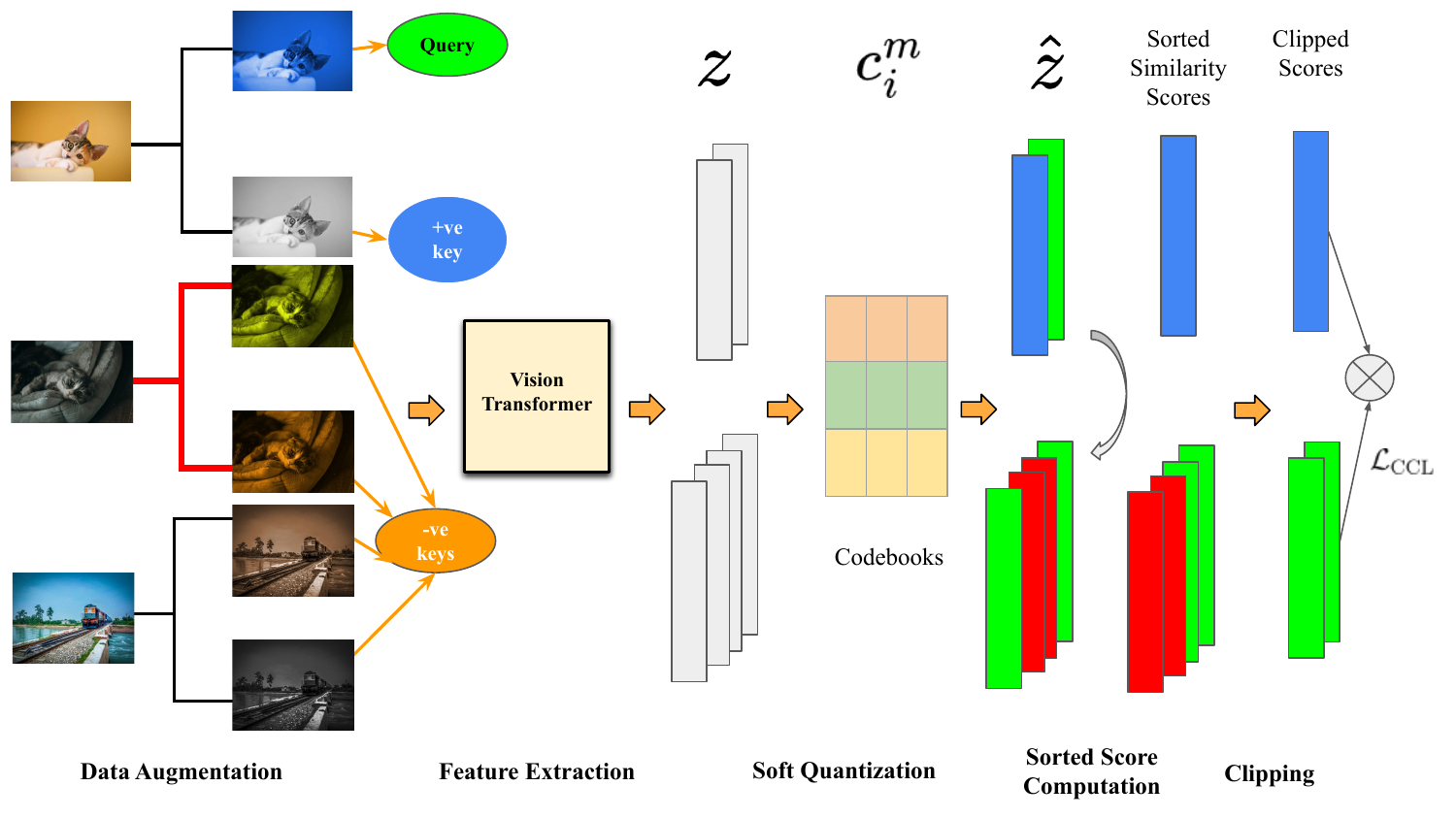}
  \caption{Schematic diagram of the proposed TransClippedCLR model. 
  }
\label{fig:arc}
\end{figure*}

The use of contrastive learning has shown promising performance for unsupervised image retrieval. Contrastive learning requires the positive and negative pairs, which are generated through the augmentated samples (i.e., different views) of a batch of images. Different augmented samples of an image form the positive pairs. However, the formation of negative pairs is non-trivial as the class label information is not available. Hence, most of the existing methods consider all pairs as the negative pairs except the two views of the same image, and suffer due to noisy false negative pairs as described in Fig. \ref{fig:motivation}.
Self-supervised Product Quantization (SPQ) \cite{spq} proposed a cross quantized learning scheme where feature vectors and soft quantized codes were compared to jointly learn codewords and deep visual descriptors. 
Contrastive Quantization with Code Memory (MeCoQ) introduced code diversity regularization and a memory bank that stores quantization codes to boost contrastive learning \cite{mecoq}. The selection of negative samples is a major problem with contrastive learning where there is high possibility that false negatives can get selected. 
Recent works have exploited different mechanism for the selection of false-negatives \cite{debiased}, \cite{park2022effective}. 
However, it leads to either increased complexity or adverse effect. We tackle these problems by clipping the potential false negative pairs.

Transformer has emerged as a prominent technique for various natural language processing applications \cite{vaswani2017attention}. Transformer network consists of stack of Transformer blocks consisting of multi-head attention (MHA) and multi-layer perceptron (MLP) modules. The MHA module utilizes the self-attention mechanism by exploiting the Query (Q), Key (K) and Value (V) projections of the input (X).
In recent years, Vision Transformer (ViT) has shown promising performance for several vision applications \cite{vit}. ViT divides the images into a grid of patches. Each patch is then treated as a sequence of tokens, and fed into the Transformer. 
CNNs employ a local approach by performing convolution within a limited area of an image. On the other hand, Visual Transformers utilize self-attention, a global technique that extracts information from the entire image/patch. This enables ViT to effectively capture semantic relationships across distant regions in an image. ViTs show promise as a more robust backend for deep hashing providing better feature represenatations than CNNs \cite{dubey2022vision}. 
The existing unsupervised methods for image retrieval utilize the CNN backbone to extract the features, which are used with different techniques to generate the hash code. The CNNs suffer to extract the global context effectively. However, the Transformer networks exploit the global context using self-attention mechanism \cite{vaswani2017attention}. In the proposed framework, we exploit the ViT \cite{vit} backbone to extract the global context.



\section{Proposed TransClippedCLR Model}

We want to learn a mapping system that functions as follows $\mathcal{S}:I \mapsto \mathbf{\hat{b}}\in\{0,1\}^\mathrm{b}$ where $\mathcal{S}$ denotes the overall system, $I$ is an image from the database of $N$ samples, and $\mathbf{\hat{b}}$ is a $\mathrm{b}$-bit binary code. The proposed TransClippedCLR model is illustrated in Fig. \ref{fig:arc}.
The first part of the system contains a deep self-attention driven transformer based feature extractor. This backbone network $\mathcal{B}(I;\theta_{\mathcal{T}})$ is responsible for taking in the image $I$ and returning a feature vector $\mathbf{z}\in R^D$, where $\theta_{\mathcal{T}}$ is the network parameter. We use ViT as backbone \cite{vit}. 
The next part is the quantization head $\mathcal{Q}(x;\theta_{\mathcal{Q}})$ which is used to transform the features into codewords. Basically, it divides the $D$ dimensional feature vector $x$ into $M$ parts. Using each of these parts $K$ clusters are defined. The parameter set for each part is called a codebook, and the clusters contained in each codebook are called codewords. Each codebook is of the form
$B_m=\{\mathbf{c}^m_{1}, \mathbf{c}^m_{2}, ..., \mathbf{c}^m_{K}\}$ where  $\mathbf{c}\in R^{D/M}$.
Codewords represent the clustered centroids of sub-vectors. These descriptors capture frequently occurring local patterns. During quantization, similar image properties are grouped together by assigning them to the same codeword, while distinct features are given different codewords \cite{spq}. 
After quantization, the distance between different views of images in a batch is used to perform the proposed clipped contrastive learning. 

\subsection{Vision Transformer Backbone}
\begin{figure}[!t]
  \centering
  \includegraphics[width=\columnwidth]{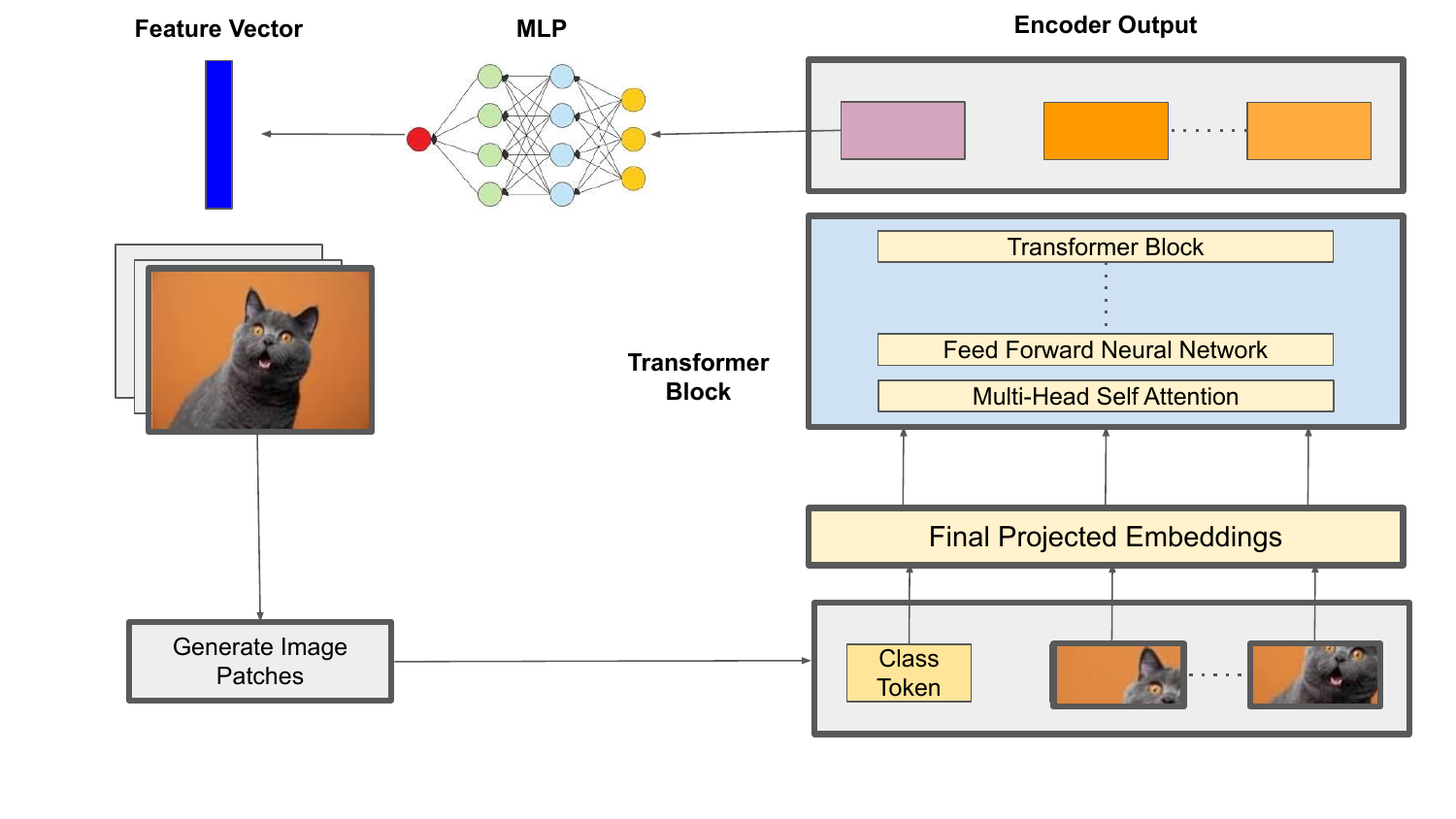}
\vspace{-8mm}
  \caption{The ViT network consisting of patch embedding, transformer encoder blocks, and multi-layer perceptron \cite{vit}. 
  }
\label{fig:vit}
\end{figure}

In this paper, we fine-tune a pre-trained Vision Transformer (ViT) backbone for feature extraction. The ViT architecture is illustrated in Fig. \ref{fig:vit}.  It consists of two main components, the Patch Embedding generator and the Transformer Encoder. Because of the quadratic attention cost, the image is divided into a number of non-overlapping patches $I_i$. 
Flattened vectors are obtained from these patches $V_i$ and then linearly projected to obtain the projected embedding as $PE_i = V_i \times W_{PE}$. The class token embedding (CT) is concatenated with these projected embedding to get the expanded embedding as $EE = [CT, PE]$. Note that the class information is used only during pre-training on ImageNet dataset. Later, it is fine-tuned on the retrieval datasets in an unsupervised manner in the proposed framework. The expanded embedding ($EE$) is used with positional embedding ($PoE$) to calculate the final projected embedding ($FE$) as $FE = dropout(EE + PoE, 0.1)$, where $dropout$ is the dropout operation \cite{dropout}.
The final projected embedding act as an input to the Transformer Encoder. The Encoder consists of a series of transformer blocks where the output of one block is the input of the next block. The basic operation of these blocks is self attention which consists of three parametric projections for Query, Key and Value vectors using the learnable weight matrices $W_Q$, $W_K$ and $W_V$, respectively. Mathematically, it can be written as $\xi_t = l_{j-1} \times W_t$, where $t \in \{Q, K, V\}$ and $l_{j-1}$ is the input feature to that encoder block.
Before computing the attention weights we calculate $\xi_{QK}$ as
$
    \xi_{QK} = \frac{\xi_Q \times \xi_K}{\sqrt{A_\xi}},
$
where $\sqrt{A_\xi}$ is the attention head size. Finally, attention weights are calculated using the softmax operation as $A_w = softmax(\xi_{QK})$.
These weights are then used with the Value vector to generate the attention weighted features as $F_A = A_w \times \xi_V$.
To improve training residual connections are applied to $F_A$. The final output of the block is obtained by passing the output of the residual block through a multilayer perceptron module followed by another residual connection. The output of the class token of the last transformer block is treated as the output feature vector of the transformer encoder.

\subsection{Quantization}
Let define the batch size as $N_B$ for a batch of images processed together. For each of these images two operators are sampled from the same data augmentation family ($T\sim\mathcal{T}$ and $T'\sim\mathcal{T}$), which are then applied to each image in the batch to obtain correlated views. More precisely, we get $2N_B$ images (views) where two separate views of the same image $(\tilde{x}_i,\tilde{x}_j)$ are correlated, and the other $2(N_B-1)$ views may comprise uncorrelated negative views having few possible false negative views. 
Consider a view of an image from a batch of images. It is first passed through the vision transformer backbone network to obtain a feature vector $\bmz\in\bbR^D$. Then, the feature vector is divided into segments $\bmz=[\bmz^1, \bmz^2, \cdots, \bmz^M]$ based on the number of codebooks used for the quantization. Let define codebooks as $\calC=[\bmC^1, \bmC^2, \cdots, \bmC^M]$, where $M$ is the number of codebooks. Each codebook consists of $K$ clusters or codewords $[\seq{\bmc^m}{K}]\in\bbR^d$, $d=D/M$. For the quantization, we compare each segment of $\bmz$ to the codewords of its codebook. If we directly assign the codeword to a feature segment, the optimization of the model becomes infeasible due to zero gradient for hard quantization assignment. Hence, we use the soft quantization during training of the model.
Each segment $\bmz^m\in\bbR^d$ of the feature vector is quantized by its corresponding codebook as $\hat{\bmz}^m =\sum_{i=1}^Kp_i^m\bmc_i^m$,
where $p_i^m$ is calcuated with the $\alpha$-softmax as,
\begin{equation}\label{equ:alpha_softmax}
    p^m_i
    =\frac{\exp\xiaok{\alpha\cdot{\bmz^m}^\top\bmc^m_i}}{\sum_{j=1}^K\exp\xiaok{\alpha\cdot{\bmz^m}^\top\bmc^m_j}},
\end{equation}
where $\hat{\bmz}^m$ is the combined score for a segment with respect to its corresponding codebook. All codewords in the codebook are utilized to approximate the quantized output, where the closest codeword contributes the most. We obtain the soft quantized reconstruction of $\bmz$ by concatenating all $\hat{\bmz}^m$ as
$\hat{\bmz}= \text{concatenate}\xiaok{\hat{\bmz}^1,\hat{\bmz}^2,\cdots,\hat{\bmz}^M}$.

\subsection{Clipped Contrastive Learning}
The soft quantized reconstructed vectors obtained from the augmented views of the same image is used as positive keys while both the views of the rest of the images in the batch are taken as negative keys. Based on this consideration, the vanilla contrastive learning (CL) loss is defined as, \begin{equation}\label{equ:vcl}
\calL_\text{VCL}=-\sum_{q=1}^{2N_B}\log\frac{\exp(\frac{S(q,k^+)}{\tau})}
{\exp(\frac{S(q,k^+)}{\tau}) + \sum_{\substack{k^-=1,\\k^-\notin\{q, k^+\}}}^{2N_B}{\exp(\frac{S(q,k^-)}{\tau})} },
\end{equation}
were $N_B$ is the batch size, $q$ is the query image to which we compare both views of all the images in a given batch of images. The positive key is represented by $k^+$ while the negative key is represented by $k^-$. $\tau$ is a non-negative temperature parameter. The distance between the query image and other images is measured using the cosine similarity function $S$.

The features that form the negative keys of the loss function come from different views of the rest of the images in the batch. As these images are randomly sampled from the dataset, there might be some images in negative pairs which are actually similar to the reference image. Due to the existence of the false negative pairs, the training of the model suffers and leads to the ambiguous feature distribution in latent space. To obviate this problem we propose a clipped contrastive loss function that provides a better approximation of the negative keys.
In the clipped contrastive loss function $\calL_\text{CCL}$, to avoid some of the false negative pairs of the vanilla loss function, we keep the numerator part of the function unchanged but in the denominator we sort the similarity scores of the negative keys in ascending order and then remove the scores which are closer to the query. The features that obtain a high similarity score when compared to the query have a high chance of coming from the similar images and thus might be false negatives and cause training problems. We define the clipped contrastive loss as, 
\begin{equation}\label{equ:ccl}
\calL_\text{CCL}=-\sum_{q=1}^{2N_B}\log\frac{\exp(\frac{S(q,k^+)}{\tau})}
{\exp(\frac{S(q,k^+)}{\tau}) + \sum_{\substack{k^-=1,\\k^-\notin\{q, k^+\}}}^{N_S - \eta}{\exp(\frac{S(q,k^-)}{\tau})} },
\end{equation}
where $N_S=2(N_B-1)$ is the number of similarity scores in the sorted similarity scores of negative keys in the increasing order of similarity. We introduce a new hyperparameter $\eta$ that controls the clipping of the scores making the total number of negative keys to $N_S - \eta$. By doing so, the negative keys having very high similarity with the query will not be considered in the loss computation. Hence, the misleading of the network due to false negative keys during training will be minimized.
Using the clipped contrastive loss ($\calL_\text{CCL}$), we can define the learning objective as
$\min_{\theta_{\mathcal{T}},\calC}\bbE\,\calL_\text{CCL}+\beta\|\theta_{\mathcal{T}}\|_2^2 + \gamma\Omega_\calC,
$
where $\Omega_\calC$ is the average inter-codeword similarity \cite{mecoq}, $\theta_{\mathcal{T}}$ denotes the network parameters of the vision transformer, $\beta$ and $\gamma$ are the trade-off hyperparameters.

\begin{table*}[!t]
    \centering
    \caption{The results of the proposed TransClippedCLR model by varying the value of clipping hyperparameter ($\eta$) in terms of mAP on CIFAR-10 (I), CIFAR-10 (II), NUS-WIDE and Flickr25K datasets. The $\eta=0$ means the vision transformer model is utilized without clipping. The best results are highlighted in \textbf{Bold}.}
    \begin{tabular}{lccclccclccclccclccc}
    \toprule
        && \multicolumn{3}{c}{CIFAR-10 (I)} &  & \multicolumn{3}{c}{CIFAR-10 (II)} &  & \multicolumn{3}{c}{NUS-WIDE} &  & \multicolumn{3}{c}{Flickr25K} \\
      \cmidrule(l){3-5} \cmidrule(l){7-9} \cmidrule(l){11-13} \cmidrule(l){15-17} 
       $\eta$ $\downarrow$ & & 16-bit & 32-bit & 64-bit & & 16-bit & 32-bit & 64-bit & & 16-bit & 32-bit & 64-bit & & 16-bit & 32-bit & 64-bit  \\
    \midrule
      0 && 87.51 & 89.74 & 90.07 && 77.55 & 90.29 & 91.04 && 79.78 & 82.64 & 83.60 && 76.69 & 78.34 & 78.24 \\ \midrule
      5 && 93.25 & 93.55 & 94.72 & & 88.70 & 94.06 & 93.89 && \textbf{81.21} & \textbf{82.88} & \textbf{84.08} && 79.20 & 77.61 & 79.19  \\
      10 & & \textbf{93.25} & 94.03 & 94.94 & & 87.99 & 93.50 & \textbf{94.93} && 81.16 & 82.87 & 83.62 && \textbf{79.58} & 78.64 & 79.72  \\
      15  && 92.10 & \textbf{94.31} & \textbf{94.94} & & 81.87 & \textbf{94.71} & 94.14 && \textbf{81.21} & 82.73 & 83.57 && 78.88 & 79.63 & 79.81 \\
      20  && 91.82 & 93.01 & 92.51 & & \textbf{88.78} & 94.23 & 93.77 && 79.71 & 82.61 & 83.18 && 78.78 & \textbf{79.95} & \textbf{80.54} \\
    \bottomrule
    \end{tabular}
    \label{tab:clipping}
\end{table*}

\subsection{Retrieval}
After contrastive learning, good representations are learnt, and a hashing function is learnt by the TransClippedCLR network to get binary hash codes. Before processing queries, the database of binary hash code is constructed from the training set using hard quantization where instead of computing the quantized features we keep track of the index of the closest codeword in the corresponding codebook of the subsegment. It can be given as $p^m_{db}=\argmax{1\le i\le K}{{\bmz^m_{db}}^\top\bmc_i^m}$.
These indices are then concatenated to form the binary code of the database image $x_{db}$. These concatenated indices form the hash codes that provide a mapping between the sub-vectors and the codewords. 
During retrieval, we similarly obtain the feature for the query image $x_q$ and divide into $M$ subsegments corresponding to the codebooks. A lookup table is set up for $x_q$ to store the precomputed similarity scores between subsegments of $x_q$ and the codewords. We adopt the asymmetric quantized similarity as the similarity metric \cite{aqs}.
Using precomputed table, we compute the asymmetric quantized similarity between the query and the database images by summing the score according to the index of the binary code.

\section{Experimental Settings}

For the experiments, three benchmark image retrieval datasets are used, including CIFAR-10 \cite{cifar}, NUS-WIDE \cite{nuswide} and Flickr25K \cite{flickr}. \textbf{CIFAR-10}~\cite{cifar}: This dataset is very widely used as the benchmark for image classification and retrieval. CIFAR-10 consists of 60k color images uniformly spread over 10 categories. The resolution of the images is $32\times32$. Following the image retrieval benchmark for CIFAR-10 dataset, we adopt two different strategies to split this dataset as CIFAR-10 (I) and CIFAR-10 (II). \emph{\textbf{CIFAR-10 (I)}}: In this setting, total 10k images form the test query set with 1k images per class. Remaining 50k images are used for training as well as to build the retrieval database.
\emph{\textbf{CIFAR-10 (II)}}: In this setting, 1k images are randomly picked as queries while 500 images per category are used to make up the training set. Retrieval database is formed by taking all of the images except the ones in the query set.
\textbf{NUS-WIDE}~\cite{nuswide}: This is a multi-label dataset with 270k images from 81 categories collected from Flickr. The 21 most popular categories are considered with 100 random images per category as the test queries. The remaining images make up the retrieval database and the training set.
\textbf{Flickr25K}~\cite{flickr}: This is also a multi-label dataset obtained from Flickr. Flickr25K dataset contains 25k images and has 24 categories. The 2k randomly picked images make up the test queries and the rest of the images are used to randomly select 5k images for the training set. All images, except the queries, are used in the retrieval database.

\begin{table*}[!t]
    \centering
    \caption{The retrieval mean Average Precision (mAP) in \% for 16, 32 and 64 bit. The performance is reported on CIFAR-10 (I), CIFAR-10 (II), NUS-WIDE and Flickr25K datasets. In class column, `D', `Q' and `BH' represent `Deep', `Quantization' and `Binary Hashing', respectively. The best and second best results are highlighted in \textbf{Bold} and \textit{Italic}, respectively.}
    \resizebox{\linewidth}{!}{
    \setlength{\tabcolsep}{.2em}{
    \begin{tabular}{lcclccclccclccclccc}
    \toprule
       & \multicolumn{2}{c}{Dataset $\rightarrow$} & &  \multicolumn{3}{c}{CIFAR-10 (I)} &  & \multicolumn{3}{c}{CIFAR-10 (II)} & \multicolumn{1}{c}{} & \multicolumn{3}{c}{NUS-WIDE} & & \multicolumn{3}{c}{Flickr25K} \\
      \cmidrule(l){2-3} \cmidrule(l){5-7} \cmidrule(l){9-11} \cmidrule(l){13-15} \cmidrule(l){17-19}
      Model $\downarrow$ & Place $\downarrow$ & Class $\downarrow$ &  & 16-bit & 32-bit & 64-bit & & 16-bit & 32-bit & 64-bit & & 16-bit & 32-bit & 64-bit & & 16-bit & 32-bit & 64-bit \\
    \midrule
      DeepBit \cite{deepbit} & CVPR'16 & DBH & & 19.43 & 24.86 & 27.73 & & 20.60 & 28.23 & 31.30 & & 39.20 & 40.30 & 42.90 &  & 62.04 & 66.54 & 68.34 \\
      Bi-halfNet \cite{bihalfnet} & AAAI'21 & DBH & & 56.10 & 57.60 & 59.50 & & 49.97 & 52.04 & 55.35 & & 76.86 & 78.31 & 79.94 &  & 76.07 & 77.93 & 78.62 \\
      CIBHash \cite{cibhash} & IJCAI'21 & DBH & & 59.39 & 63.67 & 65.16 & & 59.00 & 62.20 & 64.10 & & 79.00 & 80.70 & 81.50 &  & 77.21 & 78.43 & 79.50 \\
     SPQ \cite{spq} & ICCV'21 & DQ & & \textit{76.80} & \textit{79.30} & \textit{81.20} & & - & - & - & & 76.60 & 77.40 & 78.50 && 75.70 & 76.90 & 77.80 \\
    MeCoQ \cite{mecoq} & AAAI'22 & DQ && 68.20 & 69.74 & 71.06 && \textit{62.88} & \textit{64.09} & \textit{65.07} && \textit{80.18} & \textit{82.16} & \textit{83.24} &  & \textbf{81.31} & \textbf{81.71} & \textbf{82.68} \\
    \midrule
      
      
      TransClippedCLR & - & DQ && \textbf{93.25} & \textbf{94.31} & \textbf{94.94} && \textbf{88.78} & \textbf{94.71} & \textbf{94.93} && \textbf{81.21} & \textbf{82.88} & \textbf{84.08} &  & \textit{79.58} & \textit{79.95} & \textit{80.54} \\
    \bottomrule
    \end{tabular}}}
    \label{tab:map}
\end{table*}

\begin{table}[!t]
    \centering
    \caption{Comparison of the proposed TransClippedCLR model with transformer-based unsupervised models.}
    \resizebox{\columnwidth}{!}{
    \setlength{\tabcolsep}{.17em}{
    \begin{tabular}{lclccclccclccclccc}
    \toprule
      Model $\downarrow$ & & 16-bit & 32-bit & 64-bit & & 16-bit & 32-bit & 64-bit \\
    \midrule
    & & \multicolumn{3}{c}{CIFAR-10 (I)} & \multicolumn{1}{c}{} & \multicolumn{3}{c}{Flickr25K} \\
      \cmidrule(l){3-5} \cmidrule(l){7-9}
    MeCoQ \cite{mecoq} + ViT & & \textit{87.51} & \textit{89.74} & \textit{90.07} && \textit{76.69} & \textit{78.34} & \textit{78.24} \\
    TransClippedCLR & & \textbf{93.25} & \textbf{94.31} & \textbf{94.94} && \textbf{79.58} & \textbf{79.95} & \textbf{80.54}\\
    
    \midrule  
    & & \multicolumn{3}{c}{CIFAR-10 (II)} & \multicolumn{1}{c}{} & \multicolumn{3}{c}{NUS-WIDE} \\
    \cmidrule(l){3-5} \cmidrule(l){7-9}
    CIBHash \cite{cibhash} + ViT & & \textit{90.30} & 92.50 & 93.80 && 77.90 & 81.00 & 82.60 \\
    ViT2Hash \cite{vit2hash} & & \textbf{94.20} & \textbf{95.10} & \textbf{95.80} && 79.70 & 81.60 & 82.60 \\
    MeCoQ \cite{mecoq} + ViT & & 77.55 & 90.29 & 91.04 && \textit{79.78} & \textit{82.64} & \textit{83.60} \\
    TransClippedCLR & & 88.78 & \textit{94.71} & \textit{94.93} && \textbf{81.21} & \textbf{82.88} & \textbf{84.08}\\
    \bottomrule
    \end{tabular}}}
    \label{tab:transformer}
\end{table}

The proposed TransClippedCLR is implemented using the PyTorch library \cite{paszke2019pytorch}. We use a mix of random cropping, horizontal flipping, image graying, and randomly applied color jitter and blur for data augmentation \cite{cibhash}. Vision Transformer model with $patchsize=32$ pretrained on ImageNet dataset is used as the backbone network. Batch size is used as $128$. The training is performed for maximum $50$ epochs with early stopping. The smoothing factor of codeword assignment $\alpha$ is set to $10$ in Eq. (\ref{equ:alpha_softmax}). The number of clusters/codewords per codebook $K$ is considered as $256$, then the binary representation of the image is given by $B=M\log_2K=8M$ bit. The value of clipping hyperparameter $\eta$ is considered empirically for different data sets and different size hash codes. The experiments are performed with $\eta \in \{5, 10, 15, 20\}$. We adapt the widely used mean Average Precision (\textbf{mAP}) as the performance measure by following the previous works~\cite{ssdh,bihalfnet,cibhash}. 
Following the trend, we adapt mAP@1000 for CIFAR-10 (I) \& CIFAR-10 (II) and mAP@5000 for NUS-WIDE \& Flickr25K datasets.

\section{Experimental Results and Analysis}

\subsection{Clipping Hyperparameter Tuning and Impact}
The results in terms of the mAP in \% are summarized in Table \ref{tab:clipping} using the proposed TransClippedCLR method for different clipping hyperparameter $(\eta)$ on CIFAR-10 (I), CIFAR-10 (II), NUS-WIDE and Flickr25K datasets. The values of $\eta$ is considered as 5, 10, 15 and 20. In order to show the impact of the proposed clipped contrastive learning, we also depict the results using same backbone network without clipping (i.e., $\eta=0$) in Table \ref{tab:clipping}. It is noticed that the performance of the proposed model is significantly improved when trained with clipped contrastive learning ($\eta \in \{5, 10, 15, 20\}$) as compared to the training done with vanilla contrastive learning ($\eta=0$) in most of the cases. The T-Test scores for $\eta=5$, $\eta=10$, $\eta=15$ and $\eta=20$ w.r.t. $\eta=0$ are $0.0034$, $0.0016$, $0.0002$, and $0.0047$, respectively, which are less than $0.05$ and show that the results are statistically improved.
It is also noticed that better performance is gained using smaller and larger $\eta$ on NUS-WIDE and Flickr25K datasets, respectively. However, $\eta$ as 15 or 20 is better suited on CIFAR-10 (I) dataset. The best $\eta$ on different datasets are governed by different factors, such as similarity between different images. 

\begin{table*}[!t]
    \centering
    \caption{The mAP (\%) of TransClippedCLR model with different batch sizes (i.e., $N_B = $32, 64, and 128) by varying the value of clipping hyperparameter ($\eta$) on different datasets using 64 bit codes. The best batch wise results are highlighted in \textbf{Bold}.}
    \resizebox{\linewidth}{!}{
    \setlength{\tabcolsep}{.65em}{
    \begin{tabular}{lccclccclccclccclccc}
    \toprule
        && \multicolumn{3}{c}{CIFAR-10 (I) 64 bit} &  & \multicolumn{3}{c}{CIFAR-10 (II) 64 bit} &  & \multicolumn{3}{c}{NUS-WIDE 64 bit} &  & \multicolumn{3}{c}{Flickr25K 64 bit} \\
      \cmidrule(l){3-5} \cmidrule(l){7-9} \cmidrule(l){11-13} \cmidrule(l){15-17} 
       $\eta$ $\downarrow$ $N_B$ $\rightarrow$ & & 32 & 64 & 128 & & 32 & 64 & 128 & & 32 & 64 & 128 & & 32 & 64 & 128  \\
    \midrule
  
      5 && 83.91 & 92.78 & \textbf{94.72} & & 80.95 & 92.50 & \textbf{93.89} && 83.23 & 83.80 & \textbf{84.08} && \textbf{80.26} & 79.78 & 79.19  \\
      10 & & 66.74 & 89.85 & \textbf{94.94} & & 76.84 & 93.06 & \textbf{94.93} && 81.28 & 83.26 & \textbf{83.62} && 79.66 & \textbf{79.99} & 79.72  \\
      15  && 52.31 & 86.68 & \textbf{94.94} & & 40.69 & 92.58 & \textbf{94.14} && 77.87 & 82.52 & \textbf{83.57} && 78.51 & 79.73 & \textbf{79.81} \\
      20  && 29.26 & 78.44 & \textbf{92.51} & & 58.92 & 85.09 & \textbf{93.77} && 72.65 & 81.23 & \textbf{83.18} && 76.14 & 79.76 & \textbf{80.54} \\
    \bottomrule
    \end{tabular}}}
    \label{tab:batch}
\end{table*}

\subsection{Results Comparison}
We compare the retrieval results of the proposed TransClippedCLR method in terms of mAP in \% with the recent state-of-the-art methods in Table~\ref{tab:map}.
The TransClippedCLR leads to better results on CIFAR-10 (I), CIFAR-10 (II) and NUS-WIDE and second best results on Flickr25K dataset.
When compared against the recent state-of-the-art deep quantization approach \cite{mecoq}, the proposed method shows an increase of
$36.73\%$, $35.23\%$, $33.61\%$ on bit lengths of 16, 32 and 64, respectively, on CIFAR-10 (I) dataset;
$41.19\%$, $47.78\%$, $45.89\%$ on bit lengths of 16, 32 and 64, respectively, on CIFAR-10 (II) dataset; and
$1.28\%$, $0.98\%$, $1.01\%$ on bit lengths of 16, 32 and 64, respectively, on NUS-WIDE dataset.

The TransClippedCLR model is able to provide the comparable performance on the Flickr25K dataset as second best performing method. However, as compared with the results of $\eta=0$ in Table \ref{tab:clipping}, it is noticed that the vision transformer itself is not very suitable on Flickr25K dataset. However, the clipped contrastive learning leads to improvement on without clipping even on Flickr25K dataset.
From the experimental results, it is noticed that the proposed TransClippedCLR outperforms the existing methods with large margin on the CIFAR-10 datasets. The potential reason for the same is related to the nature of CIFAR-10 images which is similar to some categories of ImageNet dataset on which the pre-training of ViT model is done. The utilization of vision transformer and clipped contrastive learning is very beneficial for improving the retrieval performance.

\subsection{Comparison with Transformer-based Models}
We compare the results of the proposed TransClippedCLR model with transformer-based models in Table \ref{tab:transformer}. The results of TransClippedCLR are always better than MeCoQ \cite{mecoq} + ViT model on all the datasets. The proposed model outperforms CIBHash \cite{cibhash} + ViT and ViT2Hash \cite{vit2hash} models also on the NUS-WIDE dataset. The TransClippedCLR model is mostly second performer on CIFAR-10 (II) dataset after ViT2Hash model. Note that ViT2Hash model exploits the softmax classification of features which is avoided in the proposed model to make it completely unsupervised.

\subsection{Impact of Batch Size}
In order to find the relation between batch size ($N_B$) and clipping hyperparameter $\eta$, TransClippedCLR is trained with different combinations of $\eta$ and $N_B$ on different datasets. The resulting mAP (\%) values are compared for 64 bit code length with $N_B$ as 32, 64 and 128 in Table \ref{tab:batch}.
A general trend is observed, where the bigger batch size dominates and gives the best result most consistently. This is due to the limited number of scores available for contrastive loss function after clipping in small batches. The same is also evident from the observation that the performance deteriorates for small batches as the value of $\eta$ is increased resulting in excessive clipping. Hence, it is suggested to use smaller $\eta$ for small batches.

\begin{table}[!t]
    \caption{Impact of clipping on ResNet backbone.}
    \centering
    \begin{tabular}{|c|c|c|c|c|c|}
    \hline
        $\eta = 0$ & $\eta = 5$ & $\eta = 10$ & $\eta = 15$ & $\eta = 20$ \\ \hline
        68.32 & \textbf{71.08} & 68.62 & 59.87 & 56.68 \\    \hline
    \end{tabular}
    \label{tab:resnet}
\end{table}

\subsection{Impact of Clipping on ResNet Backbone}
Table \ref{tab:resnet} summarizes the mAP results for different clipping factor ($\eta$) with ResNet50 backbone on CIFAR-10 (I) dataset. The 64 bit code length is used in this experiment. The mAP is highest when $\eta = 5$ and improved by $4.04\%$ as compared to without using clipping, i.e., $\eta = 0$. It shows the generalization of the proposed clipping concept.

\section{Conclusion}
A vision transformer and clipped contrastive learning based TransClippedCLR method is proposed for unsupervised image retrieval. The vision transformer backbone is very discriminative. The proposed clipped contrastive learning discards the potential false negative keys. The TransClippedCLR outperforms the most of the existing transformer and CNN-based models. The performance by utilizing the proposed clipped contrastive learning is greatly improved as compared to the same backbone network with vanilla contrastive learning. The TransClippedCLR is better suited with large batch size. For small batches, the clipping hyperparameter is expected to be smaller. 

\section*{Acknowledgement}
This research is funded by Global Innovation and Technology Alliance (GITA) on behalf of Department of Science and Technology (DST), Govt. of India under India-Taiwan joint project GITA/DST/TWN/P-83/2019.

\bibliographystyle{IEEEbib}
\bibliography{References}

\end{document}